\documentclass{article}

\usepackage{PRIMEarxiv}

\usepackage[utf8]{inputenc} 
\usepackage[T1]{fontenc}    
\usepackage{hyperref}       
\hypersetup{
    pdfborder={0 0 0}, 
}
\usepackage{url}            
\usepackage{booktabs}       
\usepackage{amsfonts}       
\usepackage{nicefrac}       
\usepackage{microtype}      
\usepackage{lipsum}
\usepackage{fancyhdr}       
\usepackage{graphicx}       
\usepackage{arabtex}
\usepackage{arabtex}
\usepackage{enumitem}
\usepackage{amsmath}
\usepackage{utf8}
\setcode{utf8}
\usepackage{dblfloatfix}
\usepackage{tabularx} 
\usepackage{longtable}
\renewcommand{\arraystretch}{2}
\graphicspath{{media/}}     

\title{AI-VaxGuide: An Agentic RAG-Based LLM for Vaccination Decisions}

\author{
    Abdellah Zeggai, Ilyes Traikia \\
  Department of Fundamental Computing and its Applications, Faculty of NTIC \\
  Abdelhamid Mehri Constantine 02 University,
 Constantine, Algeria\\
  \textit{abdellah.zeggai@univ-constantine2.dz} \\
     \And
  Abdelhak Lakehal \\
  BIOSTIM Laboratory, Medicine Faculty \\
  Salah Boubnider Constantine 03 University, Constantine, Algeria\\
  \textit{abdelhak.lakehal@univ-constantine3.dz} \\
   \And
  Abdennour Boulesnane \\
  BIOSTIM Laboratory, Medicine Faculty \\
  Salah Boubnider Constantine 03 University, Constantine, Algeria\\
  \textit{aboulesnane@univ-constantine3.dz} \\
}
\renewenvironment{abstract}
{
  \small    
  \begin{center}
  \bfseries Abstract  
  \end{center}
  \begin{quote}       
}
{
  \end{quote}
}

\begin{document}
\maketitle

\begin{abstract}
Vaccination plays a vital role in global public health, yet healthcare professionals often struggle to access immunization guidelines quickly and efficiently. National protocols and WHO recommendations are typically extensive and complex, making it difficult to extract precise information, especially during urgent situations. This project tackles that issue by developing a multilingual, intelligent question-answering system that transforms static vaccination guidelines into an interactive and user-friendly knowledge base. Built on a Retrieval-Augmented Generation (RAG) framework and enhanced with agent-based reasoning (Agentic RAG), the system provides accurate, context-sensitive answers to complex medical queries. Evaluation shows that Agentic RAG outperforms traditional methods, particularly in addressing multi-step or ambiguous questions. To support clinical use, the system is integrated into a mobile application designed for real-time, point-of-care access to essential vaccine information. AI-VaxGuide model is publicly available on \url{https://huggingface.co/VaxGuide}
\end{abstract}


\section{Introduction}
Vaccination is widely recognized as one of the most effective public health interventions, having played a pivotal role in reducing the global burden of infectious diseases and saving millions of lives annually. Historical milestones such as the eradication of smallpox and the substantial decline in diseases like polio, measles, and diphtheria underscore the critical importance of vaccines in disease prevention and population health management~\cite{who2024immunization, greenwood2014vaccination}.
Effective immunization strategies depend heavily on the timely dissemination and utilization of clinical guidelines issued by organizations such as the World Health Organization (WHO) and national health ministries. These documents, however, are typically provided in static formats such as PDFs or websites, which require healthcare professionals to extract the relevant information manually. This process is often inefficient, especially in high-pressure clinical or emergency settings where rapid decision-making is essential.

Recent advancements in Artificial Intelligence (AI), particularly through large language models (LLMs), have shown promising potential across multiple healthcare domains. From early disease detection and medical imaging to personalized treatment planning and health education, AI-driven solutions are transforming the delivery of care~\cite{lancet2025ai, tornimbene2025ai}. In particular, conversational AI systems are increasingly utilized to assist clinicians and patients by providing timely, context-aware responses to medical inquiries.

Despite these advancements, the application of AI in immunization-related tasks remains underdeveloped compared to other medical fields such as oncology or radiology. The current gap presents a compelling opportunity to explore how LLM-based systems can enhance access to vaccination information, improve workflow efficiency, and support informed clinical decision-making in immunization practices.

In this paper, we propose an intelligent, multilingual question-answering system built upon a Retrieval-Augmented Generation (RAG) architecture tailored to the domain of vaccination. Our system transforms static immunization guidelines, including the Algerian national vaccination protocol and WHO documents, into an interactive, searchable knowledge base. We further enhance the system using an Agentic RAG framework that enables multi-step reasoning across complex, structured documents. To validate our approach, we conduct a comprehensive evaluation using domain-specific metrics and deploy the solution within a user-friendly mobile application designed for healthcare professionals.

The rest of the paper is organized as follows: Section \ref{sec3} provides an overview of related work on LLMs in Vaccination. Section \ref{sec4} introduces the proposed methodology. Section \ref{sec5} evaluates and discusses the experimental findings. Finally, Section \ref{sec6} concludes the paper and suggests future research directions.

\section{Related Work on LLMs in Vaccination}\label{sec3}
Large Language Models (LLMs) are increasingly explored for diverse applications within the vaccination domain, spanning safety monitoring, public communication, hesitancy analysis, clinical workflows, and even vaccine development. This section reviews key research efforts in these areas.

Several studies have leveraged LLMs to automate the extraction and analysis of adverse events (AEs) related to vaccination from unstructured sources. For instance, \cite{Wang2025} demonstrated the effectiveness of LLMs in extracting AE data from vaccine package inserts and using text embeddings to distinguish between live and non-live vaccines with high accuracy. Similarly, \cite{Li2024} fine-tuned GPT-3.5 ("AE-GPT") to extract AEs from the U.S. Vaccine Adverse Event Reporting System, achieving strong results for influenza vaccines. In another approach, \cite{Abate2025} evaluated off-the-shelf models like ChatGPT and Gemini for causality assessment of AEs following COVID-19 immunizations. While ChatGPT showed better adherence to the WHO algorithm and moderate agreement with human experts, limitations remained, suggesting LLMs should be viewed as complementary rather than standalone tools.

LLMs are also being investigated for their role in public health communication and combating misinformation. \cite{Deiana2023} assessed ChatGPT's ability to respond to WHO vaccination myths, reporting 85.4\% accuracy with improved performance in GPT-4.0, though misinterpretation risks and ethical concerns with paid versions were noted. \cite{Sohail2023} found that ChatGPT's supportive responses could reduce misconceptions and shape positive perceptions. \cite{Joshi2024}, reporting on the same study, compared ChatGPT with CDC vaccine information in English and Spanish, highlighting generally high accuracy and understandability but also issues with readability and Spanish translation quality, emphasizing the need for linguistic equity. \cite{Gullison09857} demonstrated that ChatGPT-augmented vaccine messages were rated as slightly more persuasive than original public health messages, suggesting potential for enhancing communication.

In monitoring public sentiment and vaccine hesitancy, LLMs have proven effective at analyzing social media discourse. \cite{Espinosa2024} showed that few-shot prompting of top LLMs outperformed traditional methods in classifying public stances toward vaccination. \cite{Liu2025} and \cite{Liu2024} analyzed a decade of Japanese tweets about HPV vaccines using Gemini 1.0 Pro, tracking stance changes linked to policy shifts and identifying trends in misinformation. Similarly, \cite{Espinosa2025} used GPT-4 to analyze Brazilian Twitter data (2013–2019), revealing a decline in positive sentiment coinciding with a measles resurgence. \cite{Sun2411.14720} optimized in-context learning strategies for stance detection, showing better performance than fine-tuning approaches. \cite{QinyuZhu2024} employed a zero-shot, multi-pass GPT-4 pipeline to hierarchically classify vaccine concerns with high accuracy, whereas \cite{De231210626} found GPT-3.5 to be most effective for multi-label concern classification. \cite{Kim2024} demonstrated that ChatGPT could accurately classify pro- and anti-HPV vaccine messages, although it struggled with nuanced, long-form pro-vaccine content. \cite{VanNooten2023} further used GPT-3.5 to generate Dutch anti-vaccine tweets for data augmentation, enhancing the classification of rare hesitancy arguments.

LLMs have also been explored in ethical, policy, and behavioral simulation contexts. \cite{Mugu2024} examined how LLMs reason about vaccination mandates using the Jacobson v. Massachusetts case, finding high alignment with expert judgment through advanced prompting, but also model variability and hallucination risks that require human oversight. \cite{Hou250309639} introduced "VacSim", a generative agent framework that simulates societal responses to vaccine policies using LLMs, revealing both the promise and alignment challenges. \cite{Sehgal250420519} found, in a multi-country randomized controlled trial, that chatbot-based conversations increased short-term HPV vaccination intent among hesitant parents. However, they did not outperform standard health materials and lacked a long-term impact, raising questions about their added value.

In clinical and operational settings, LLMs are assisting healthcare workflows. \cite{Mulleners2025} used LLMs to efficiently screen literature for a review on vaccine safety and efficacy during lactation. \cite{Cosma2025} piloted the use of ChatGPT to enhance the readability and understandability of vaccination consent forms, with best results achieved when human oversight was included. \cite{Burstein2024} applied various LLM strategies (zero-shot, few-shot, fine-tuning) to categorize reasons for non-vaccination in survey responses from the Democratic Republic of the Congo, achieving high accuracy with minimal examples.

Finally, LLMs are also being explored for vaccine research and development. \cite{Li2024b} introduced "CodonBERT", an LLM tailored to optimize mRNA sequences using codon inputs, outperforming prior methods on flu vaccine prediction tasks. \cite{Ahmad2503.09103} and \cite{Hayawi2024}, sharing identical abstracts, discussed the role of generative AI and LLMs in reverse vaccinology, highlighting improvements in candidate identification but also raising ethical concerns regarding bias and data consent.
\section{Methodology and Proposed Approaches}\label{sec4}
The initial design of our system followed a conventional RAG pipeline architecture. Subsequently, the system was extended with agent-based functionalities to address complex queries more effectively, facilitate document-specific reasoning, and enable context-aware dynamic tool utilization.

The fundamental objective remains consistent: to enhance the capabilities of a large language model (LLM) by leveraging external knowledge stored in a vector database. An agentic layer is then incorporated, which decomposes the document corpus into modular tools. Each tool corresponds to a specific section or individual document within the vaccination guides, thereby enabling more targeted and accurate retrieval.

This work presents a comprehensive overview of the complete pipeline and methodology, encompassing document parsing and preprocessing, embedding generation, indexing, retrieval, and the question-answering mechanism supported by agentic orchestration.
\begin{figure}[t]
    \centering
    \includegraphics[width=0.95\textwidth]{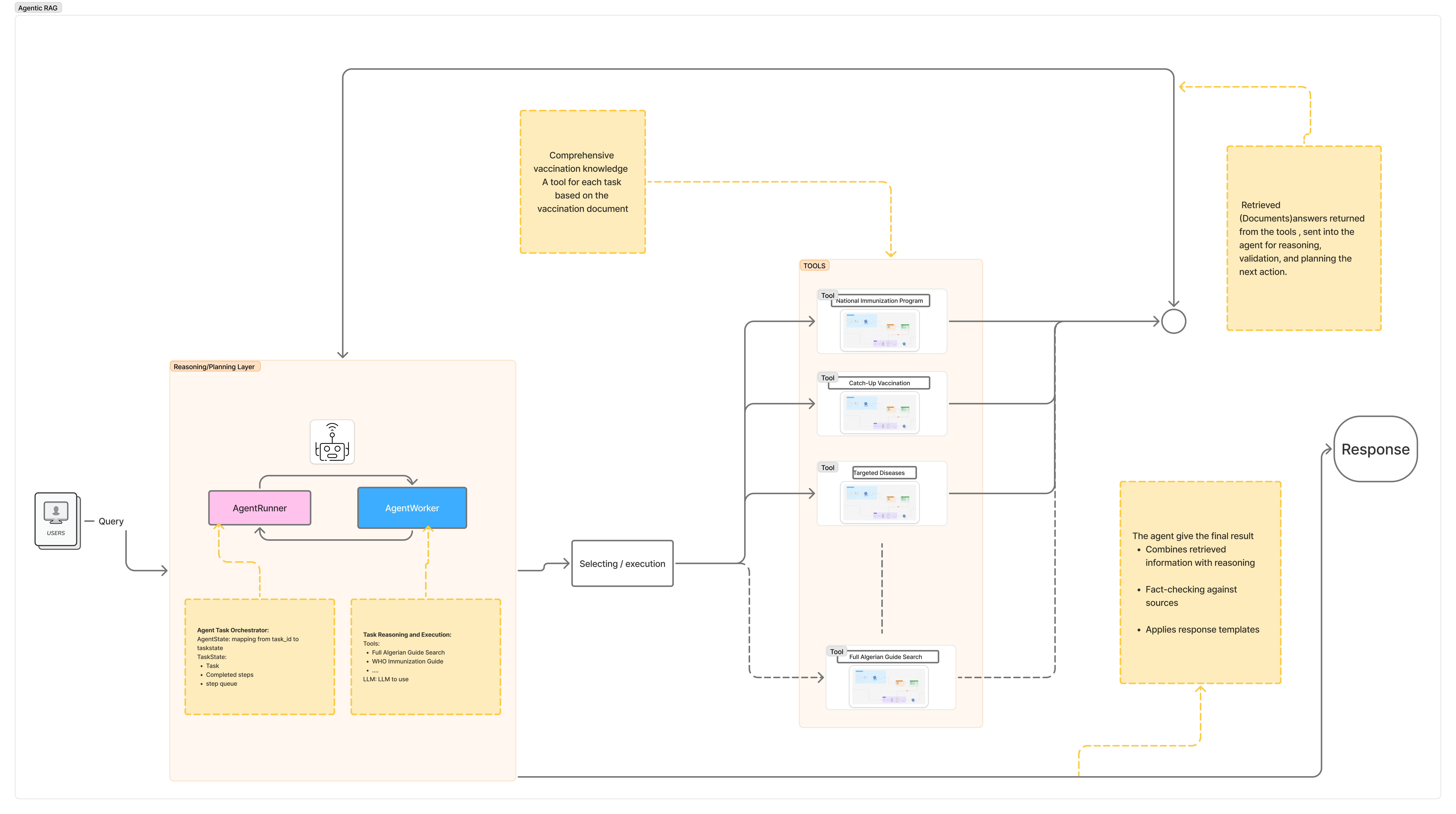}
    \caption{Agentic RAG.}
    \label{fig:Agentic}
\end{figure}
Figure~\ref{fig:Agentic} depicts the evolved agent-based architecture. However, it is essential first to understand the underlying RAG pipeline architecture, which is embedded within each tool. Each section-specific tool employed by the agent operates on the same foundational RAG process: content-specific preprocessing, embedding into a vector space, retrieval of relevant segments based on user queries, and response generation via the LLM.
In essence, the agentic system comprises multiple self-contained RAG pipelines, with each pipeline dedicated to a specific document or section. This modular structure allows the agent to reason about and select the most relevant section(s) for responding to a given query, thereby improving accuracy and contextual relevance.

This next figure (see Figuer~\ref{fig:RAGSimple}) is the baseline Vector-based RAG system that we will first explain in detail:
\begin{figure}[h!]
    \centering
    \includegraphics[width=0.95\textwidth]{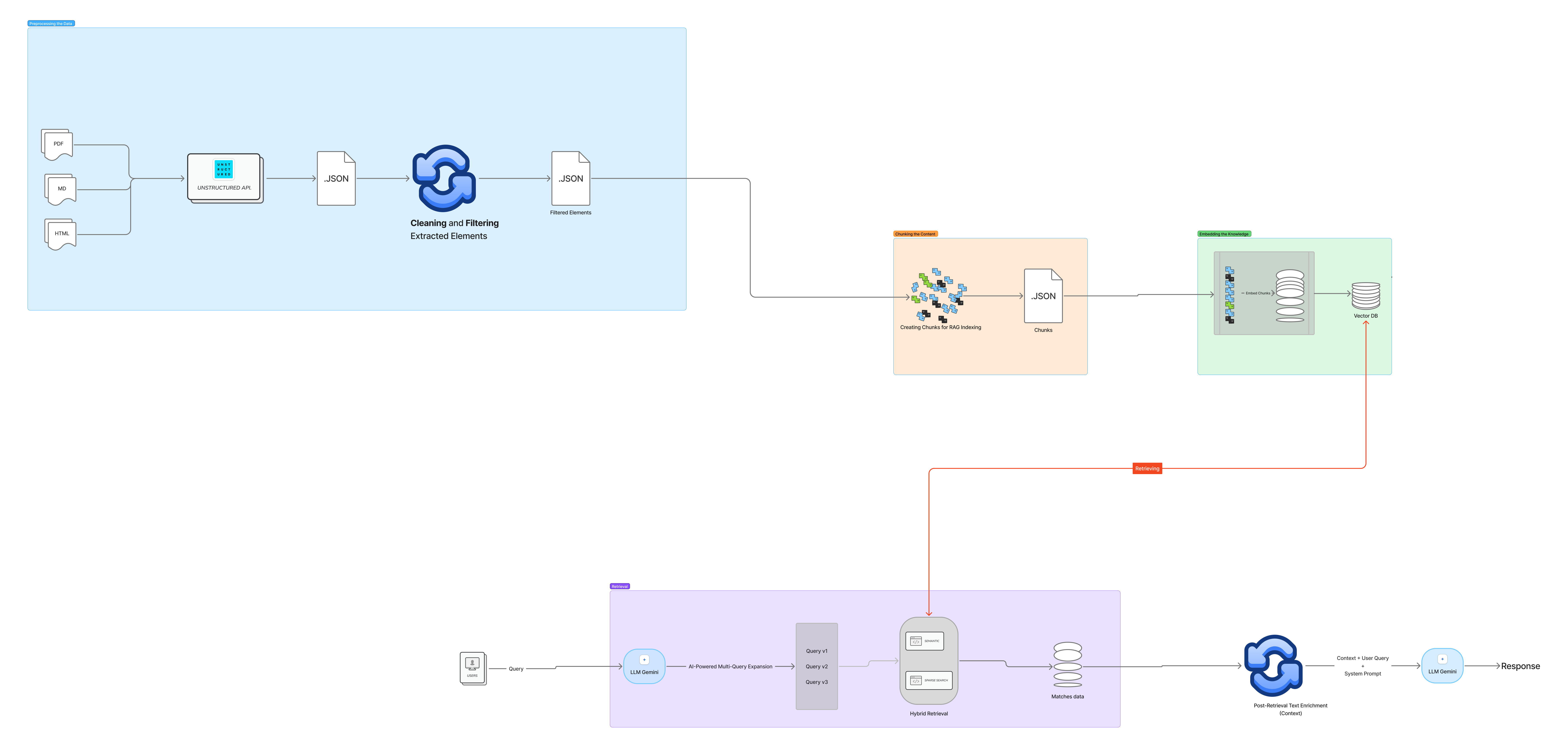}
    \caption{Simple RAG.}
    \label{fig:RAGSimple}
\end{figure}

The Libraries used for the system are: \textit{unstructured--client}, \textit{langchain\_google\_genai}, \textit{langchain\_community}, \textit{langchain\_core}, \textit{chromadb}, \textit{rank\_bm25}.
\subsection{Data Preprocessing}
The preprocessing stage is essential, as raw PDF documents typically contain a heterogeneous mixture of narrative text, tables, and complex formatting that is not directly compatible with language model inputs. This step transforms the raw content into semantically coherent components—such as \textbf{NarrativeText} and \textbf{Title}—that are better suited for RAG tasks.

To achieve this, we employed the Unstructured Framework API for parsing the vaccination documents. The initial step involves partitioning the PDF content specifically for integration into the RAG pipeline. Using the Unstructured API, each PDF file is loaded and processed to extract structured representations of its content, which are subsequently serialized into JSON format for downstream processing.
\subsection{Cleaning and Filtering Extracted Elements}
The second stage involves cleaning and filtering the extracted elements to prepare them for indexing. Certain portions of the document, such as footnotes, tables of contents, and boilerplate administrative content, do not contribute meaningful information for question answering (QA). Eliminating these sections improves retrieval quality and reduces the likelihood of generating misleading responses.

Additionally, we identify and extract unique element types (e.g., \textit{NarrativeText}, \textit{Table}, \textit{Title}, and \textit{UncategorizedText}) to better understand the document structure. To further refine the dataset, we remove all elements located between two specific section titles that typically delimit non-informative content. This ensures that only semantically relevant portions are retained for embedding and subsequent retrieval in the RAG system.

\begin{figure}[h!]
\centering
\includegraphics[scale=0.1]{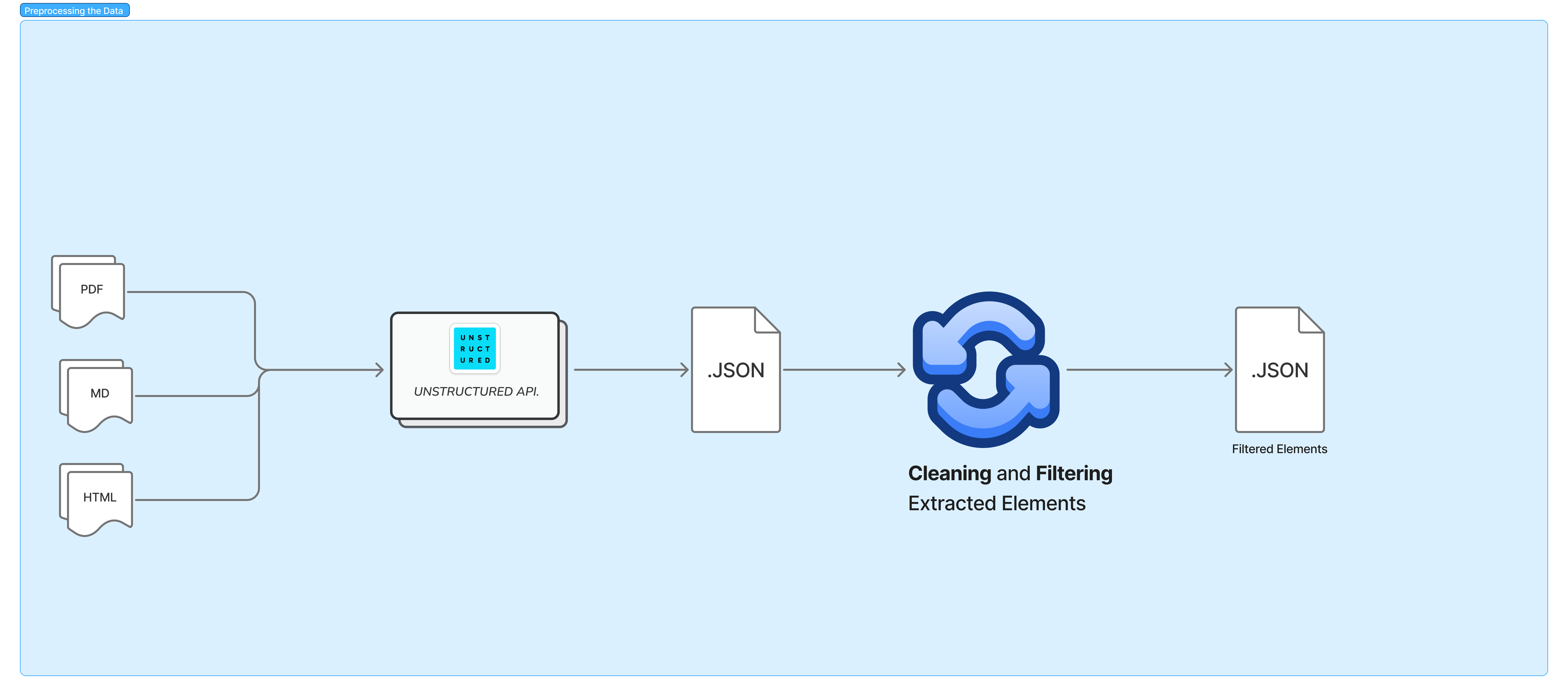}
\caption{Data preprocessing and cleaning pipeline.}
\label{fig:AbdeallahRAG}
\end{figure}

The effect of this cleaning step on QA quality is illustrated in Table~\ref{tab:cleaning_effect}, where we compare representative examples of document chunks and model responses before and after the cleaning process.
\begin{table}[h!]
\caption{Impact of Data Cleaning on Response Quality}
\label{tab:cleaning_effect}
\centering
\renewcommand{\arraystretch}{1.2}
\begin{tabular}{p{0.45\textwidth} p{0.45\textwidth}}
\toprule
\textbf{Before Cleaning} & \textbf{After Cleaning} \\
\midrule
\textit{Chunk:} "Ministère de la santé – Table des matières – Page 1..." \newline
\textit{Response:} "Je suis désolé, je n’ai pas pu trouver cette information." &
\textit{Chunk:} "La vaccination de rattrapage doit respecter l’âge et les intervalles entre les doses..." \newline
\textit{Response:} "Selon la règle 8 du rattrapage vaccinal, l’âge minimal doit être respecté entre les doses." \\
\bottomrule
\end{tabular}
\end{table}
Figure~\ref{fig:AbdeallahRAG} illustrates this preprocessing workflow and highlights the data cleaning stage within the broader system pipeline.

\subsection{Semantic Chunking}
\begin{figure}[h!]
\centering
\includegraphics[scale=0.25]{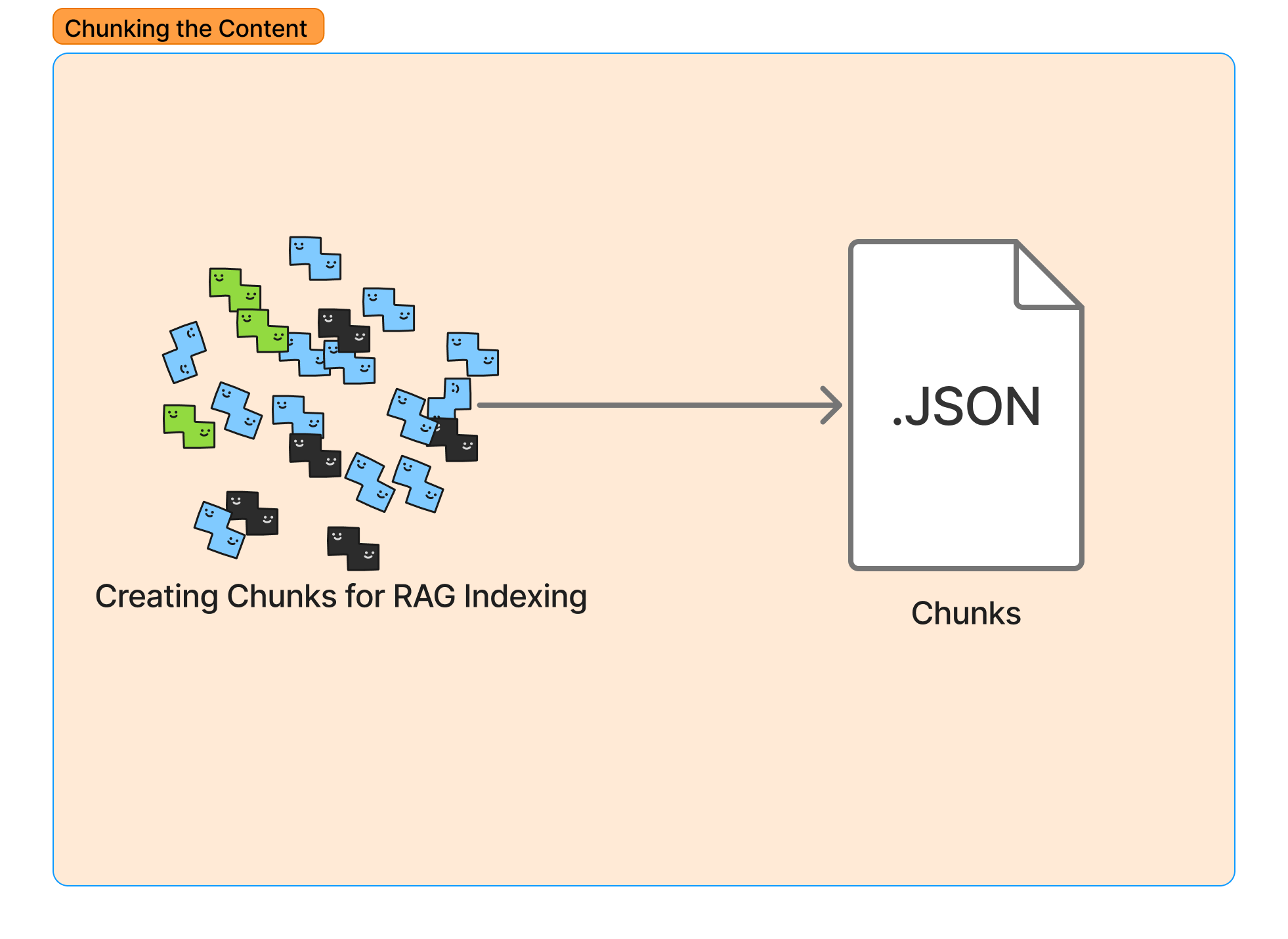}
\caption{Semantic chunking based on document structure.}
\label{fig:Embeddings}
\end{figure}

The subsequent step in the pipeline is semantic chunking, which is a critical component for optimizing RAG. Unlike arbitrary segmentation methods (e.g., fixed-length token or character-based splits), semantic chunking respects the inherent structure of the document by grouping content under meaningful headings. Specifically, content is segmented based on title elements, allowing for the creation of thematically coherent units suitable for RAG indexing, as illustrated in Figure~\ref{fig:Embeddings}.

When a table is encountered within a chunk, it is processed separately using specialized logic. Each table is enriched with its original content, an optional AI-generated French description, and an HTML representation. These enriched tables are stored as \textbf{TableElement} entries, each assigned a unique \textit{element\_id}.

Tables are handled independently due to their format, which can be challenging for traditional retrievers to index and interpret, especially for structured content such as vaccination schedules. By enriching them with a descriptive summary, their discoverability and relevance in downstream retrieval tasks are significantly enhanced.

Each chunk is preserved in its original logical grouping and stored as a \textbf{CompositeElement} object. This includes a unique \textit{element\_id}, combined textual content, metadata (e.g., filename and file type), and the list of constituent elements. All generated chunks—including both composite and table elements—are serialized and saved into a structured JSON file (\textit{chunks.json}) for use in downstream indexing and retrieval stages.

This chunking strategy ensures that content is grouped in a contextually meaningful manner, thereby facilitating more accurate and context-aware responses during question answering or summarization tasks. The effect of semantic chunking on QA performance is illustrated in Table~\ref{tab:chunking_example}.
\begin{table}[h!]
\caption{Impact of Semantic Chunking on Response Quality}
\label{tab:chunking_example}
\centering
\renewcommand{\arraystretch}{1.2}
\begin{tabular}{p{0.45\textwidth} p{0.45\textwidth}}
\toprule
\textbf{Before Semantic Chunking} & \textbf{After Semantic Chunking} \\
\midrule
\textit{Question:} Quelle est le Calendrier vaccinal 2023 de vaccination ? \newline
\textit{Response:} Désolé, je ne sais pas. &
\textit{Question:} Quelle est le Calendrier vaccinal 2023 de vaccination ? \newline
\textit{Response:} Le Calendrier national 2023 consiste en des vaccins tels que BCG, HBV à la naissance... \\
\bottomrule
\end{tabular}
\end{table}

\subsection{Chunk Embedding and Storage}
\begin{figure}[h!]
\centering
\includegraphics[scale=0.25]{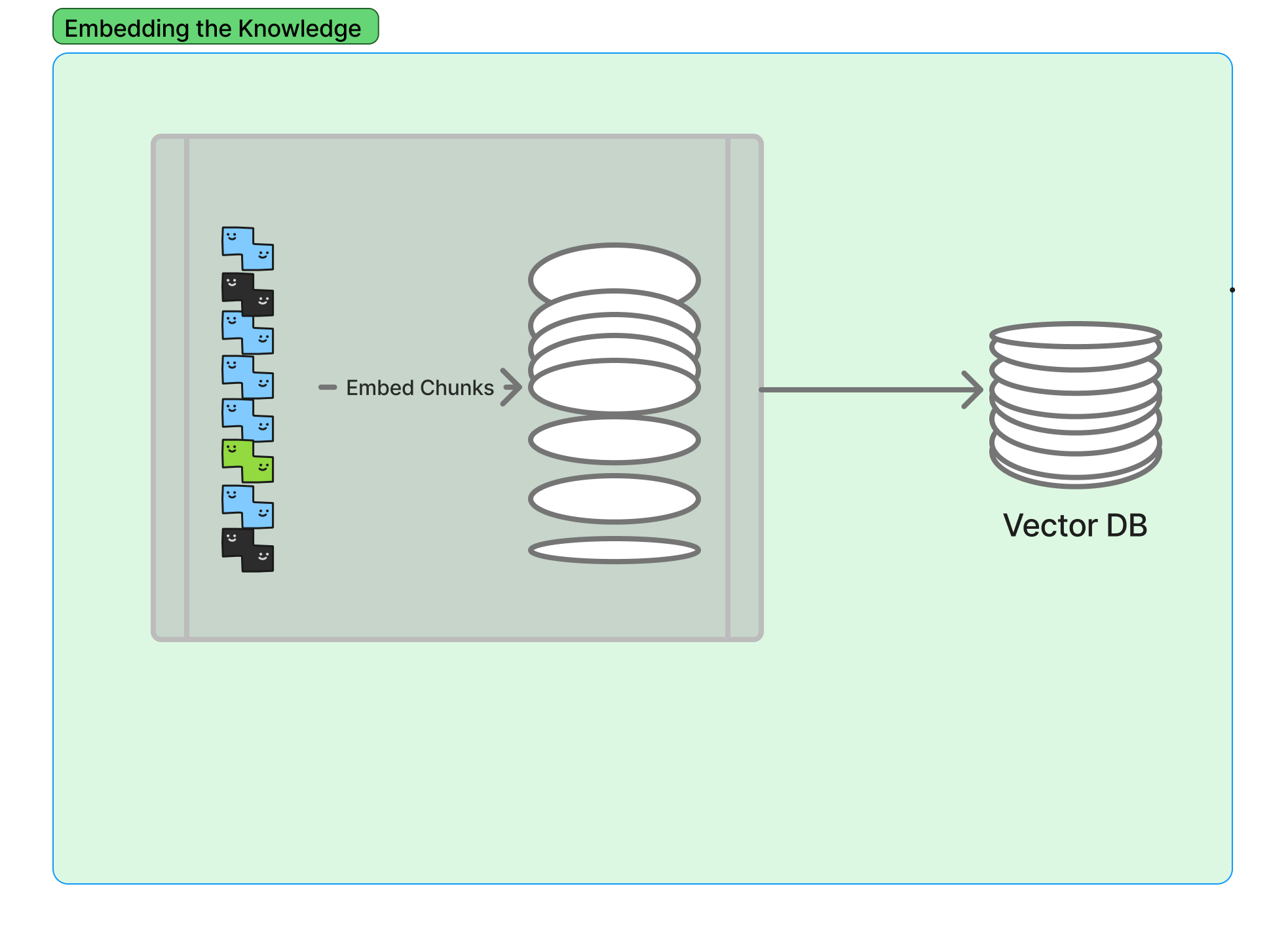}
\caption{Embedding and storing document chunks for semantic retrieval.}
\label{fig:EmbeddingAbdellah}
\end{figure}

To enable efficient semantic retrieval, the cleaned and semantically grouped chunks must be transformed into a vector representation suitable for similarity search. This step, illustrated in Figure~\ref{fig:EmbeddingAbdellah}, involves embedding the chunks and storing them in a vector database, forming the foundation of the RAG system.

Each chunk, extracted from the \textit{chunks.json} file, is converted into a LangChain \textit{Document} object. Associated metadata, such as language, file name, and file type, is appended to each document. For table-based chunks, the HTML representation of the table is also included in the metadata to preserve structural information and enhance retrievability.
For vector embedding, we employ the \textit{intfloat/multilingual-e5-base} model via the HuggingFace \textit{Embeddings} interface. This multilingual model is particularly well-suited for processing documents in French and Arabic, aligning with the linguistic characteristics of the vaccination guide.

The resulting embedded vectors are stored using ChromaDB, a lightweight and high-performance vector database. A dedicated collection, labeled \textit{Guide\_2023\_e5\_multilingual}, is created to group all embedded documents, and the database is saved locally in the directory \textit{chroma\_db\_multilingual}.

This embedding and storage process enables fast and semantically rich retrieval, allowing the downstream agent-based components to effectively match user queries with relevant content from the vaccination documents. It forms a critical layer of the RAG pipeline, supporting intelligent and context-aware question answering.

\subsection{Hybrid Multi-Retriever Strategies}

To enhance retrieval quality and ensure robust coverage of both semantically rich and keyword-specific user queries, we implement a hybrid retrieval framework that integrates multiple complementary strategies. This design improves the accuracy and resilience of the RAG pipeline, particularly when handling ambiguous, paraphrased, or domain-specific questions.

\textbf{Multilingual Vector Retriever:}
A dense semantic retriever based on the \textit{intfloat/multilingual-e5-base} model is employed to support French and Arabic queries. It enables the system to understand contextually similar but lexically diverse user inputs, including paraphrased or imprecise formulations.

\textbf{Sparse BM25 Retriever:}
To compensate for the limitations of dense retrievers in handling rare or technical terms, we incorporate a keyword-based BM25 retriever. This mechanism is particularly effective for retrieving passages containing exact matches, which is crucial for terms such as "rattrapage."

\textbf{Ensemble Retriever:}
By combining semantic and sparse retrievers in an ensemble configuration, we balance contextual understanding with lexical precision. The two retrievers are weighted equally, resulting in improved recall and robustness across diverse query types.

\textbf{LLM-Powered Multi-Query Expansion:}
User queries are often vague or under-specified. To address this, we employ a large language model (Gemini 2.0 Flash) to generate multiple semantically diverse reformulations of the original query. This strategy significantly expands the search space and increases the likelihood of retrieving relevant content, particularly for layperson-formulated questions.
\begin{figure}[h!]
\centering
\includegraphics[width=0.8\textwidth]{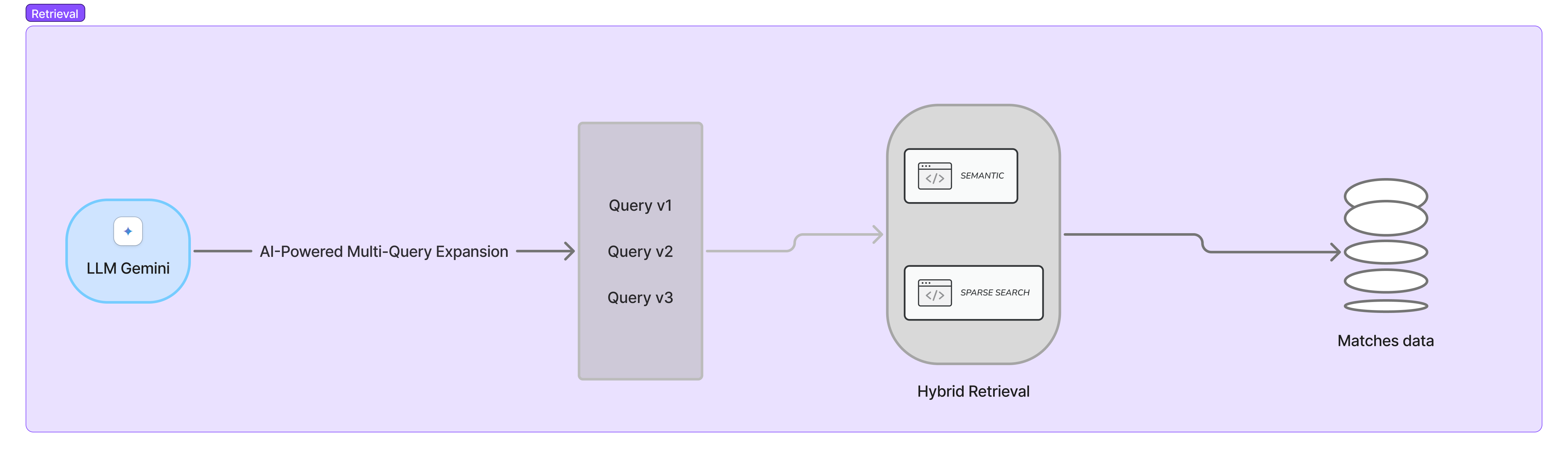}
\caption{Hybrid multi-retriever architecture and query expansion process.}
\label{fig:Retrival}
\end{figure}

The fourth stage of the pipeline, illustrated in Figure~\ref{fig:Retrival}, consists of the hybrid retrieval system and query expansion module. The configuration of each component is detailed below:

\begin{enumerate}[label=(\alph*)]
\item \textit{Multilingual Vector Retriever:} The embedded vector store is used to construct a semantic retriever, returning the top $k=6$ chunks most similar to the user query, using multilingual embeddings.

\item \textit{Sparse BM25 Retriever:} A keyword-based retriever based on the BM25 algorithm returns the top $k=2$ results based on exact token matches and term frequency.

\item \textit{Ensemble Retriever:} Both the semantic and sparse retrievers are combined using the \textit{EnsembleRetriever}, with equal weighting ($0.5$ each) to balance recall and precision.

\item \textit{LLM-Based Query Expansion:} The original query is expanded using an LLM-based \textit{MultiQueryRetriever} (via ChatGoogleGenerativeAI). This generates multiple alternative queries, increasing the breadth of retrieval coverage.
\end{enumerate}

This hybrid retrieval architecture ensures a comprehensive response generation capability, capturing both semantic meaning and lexical specificity. Table~\ref{tab:hybrid_retrieval} demonstrates the improvement in answer quality when using hybrid retrieval and query expansion versus basic retrieval.

\begin{table}[h!]
\caption{Effect of Hybrid Retrieval and Query Expansion on Response Quality}
\label{tab:hybrid_retrieval}
\centering
\renewcommand{\arraystretch}{1.2}
\begin{tabular}{p{0.45\textwidth} p{0.45\textwidth}}
\toprule
\textbf{Without Expansion / Basic Retrieval} & \textbf{With Hybrid + Multi-Query Expansion} \\
\midrule
\textit{Question:} Quelle est la règle 8 du rattrapage ? \newline
\textit{Response:} Désolé, je ne trouve pas de règle correspondante. &
\textit{Question:} Quelle est la règle 8 du rattrapage ? \newline
\textit{Response:} La règle 8 du rattrapage stipule que l'on doit administrer le vaccin DTC au minimum 4 semaines après la dose précédente. \\
\bottomrule
\end{tabular}
\end{table}
\subsection{RAG Pipeline Execution}
The final phase of the system involves executing the RAG pipeline, which integrates Google's Gemini Flash 2.0 model with carefully designed prompt engineering techniques.
At the core of this stage lies the \textit{rag()} function, which orchestrates three key components: \textbf{prompt formulation}, \textbf{context assembly}, and \textbf{response generation} using the Gemini Flash 2.0 language model.

The \textit{rag()} function is tailored for domain-specific applications, particularly in medical and vaccination-related contexts. It explicitly frames Gemini as a virtual assistant for healthcare professionals, constrains its responses strictly to the retrieved content, and ensures the proper formatting of structured data, such as HTML tables, by rendering them in Markdown while preserving their original layout.
To further enhance retrieval accuracy and coverage, the system employs the \textit{expanding\_retriever} mechanism. This approach leverages Gemini to reformulate user queries, enabling broader and more effective document retrieval from both the multilingual vector retriever and the sparse BM25 retriever. If available, relevant HTML table content is appended to the final context.

\begin{figure}[h!]
\centering
\includegraphics[width=0.95\textwidth]{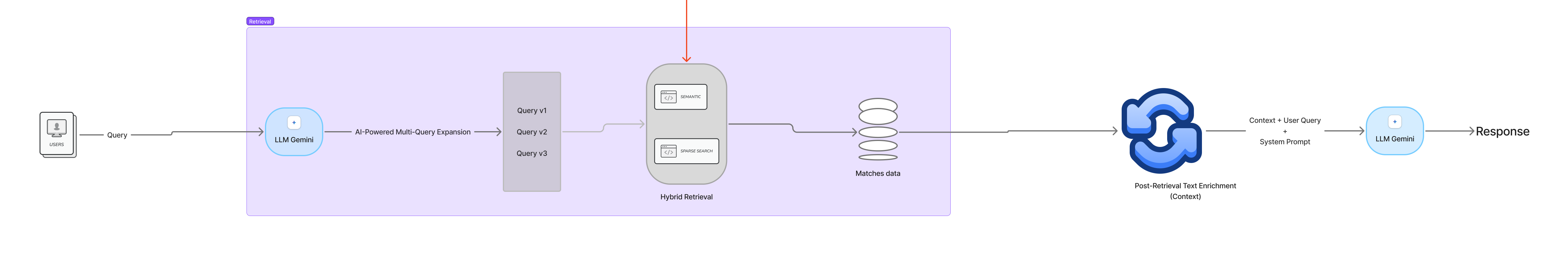}
\caption{Final step of the system: Execution of the RAG pipeline with prompt engineering and Gemini integration.}
\label{fig:FinalStep}
\end{figure}

The overall process is illustrated in Figure~\ref{fig:FinalStep}, which depicts the architecture and flow of the RAG execution stage.

\subsection{Agentic Layer and Vector Tools}
To enhance the system’s reasoning capabilities and enable it to handle complex medical queries, an agent-based layer was introduced on top of the base RAG pipeline. This agentic architecture enables the system not only to retrieve relevant information but also to perform multi-step reasoning and structured decision-making before generating a response.
\begin{figure}[h!]
\centering
\includegraphics[scale=0.22]{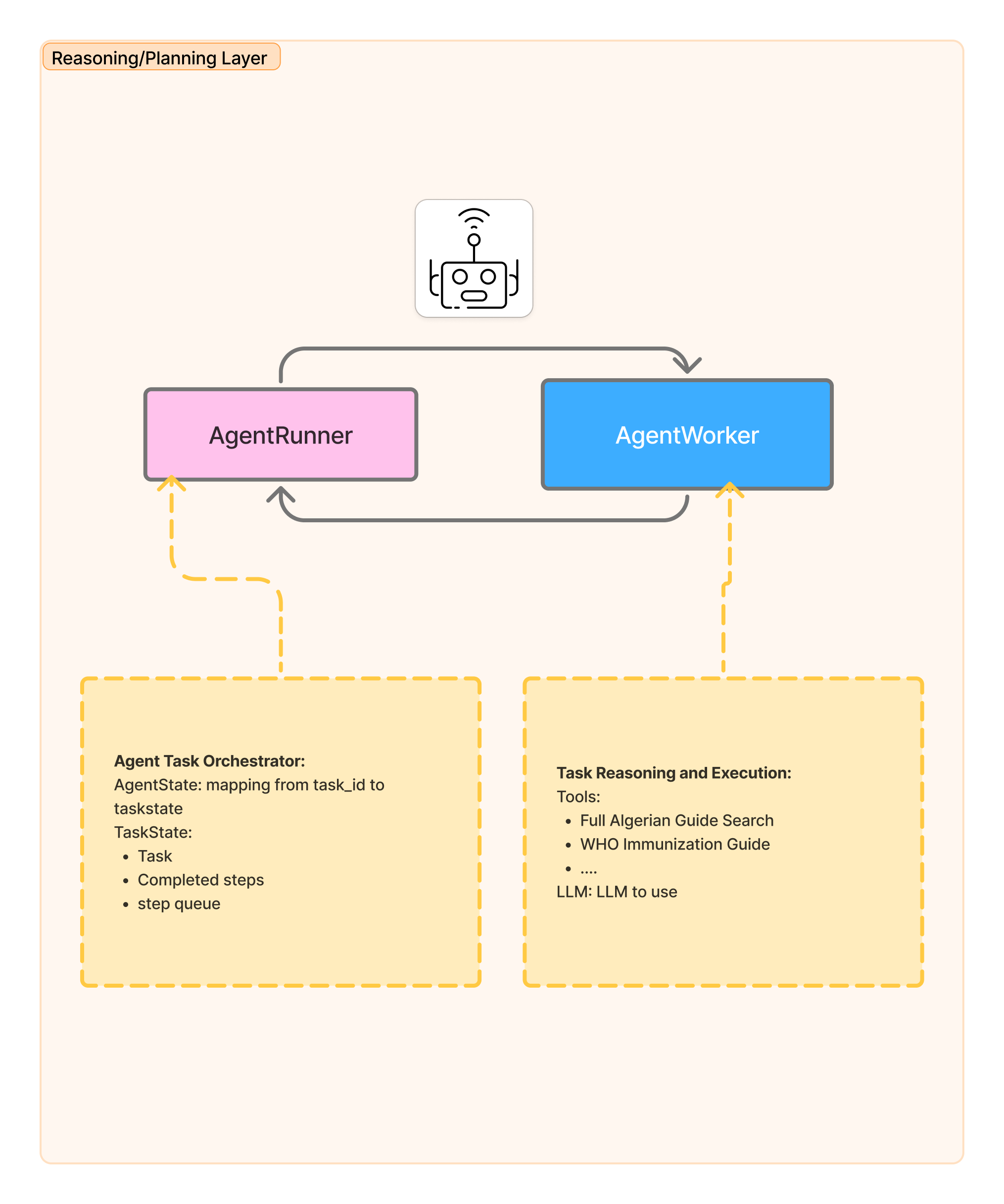}
\caption{Planning phase within the agentic framework.}
\label{fig:plannig}
\end{figure}

\textbf{Planning and Task Orchestration:}
The first stage involves a planning mechanism coordinated by a central controller agent. This agent decomposes incoming queries into a sequence of subtasks, manages task dependencies, and tracks execution progress. Inspired by the ReAct paradigm~\cite{yao2023reactsynergizingreasoningacting}, the system maintains a memory of completed tasks, enabling efficient orchestration and avoiding redundant processing.

\textbf{Tool Selection and Execution:}
Following the planning phase, the agent selects appropriate tools based on the task requirements. Each tool corresponds to a specific document or section, such as chapters from the Algerian national vaccination guidelines or WHO publications, and is accompanied by a metadata description. These descriptions help the agent align user intent with the relevant knowledge source.

During execution, the agent invokes only the tools deemed relevant by the reasoning strategy. For instance, a query concerning neonatal immunization may prompt the agent to select the module linked to the corresponding section of the national guide. This targeted retrieval approach enhances both precision and efficiency.

\textbf{Final Reasoning and Response Generation:}
In the final stage, the agent aggregates the outputs retrieved by the selected tools and performs a global reasoning pass. It validates the information for coherence and factual correctness, cross-checking it against established sources~\cite{yao2023reactsynergizingreasoningacting}.

Upon validation, the system generates a response using pre-defined templates to ensure clarity, fluency, and consistency. This stage integrates the understanding of user intent, tool selection, evidence aggregation, and response synthesis, thereby enabling robust, explainable answers grounded in domain-specific medical knowledge.

\section{Experimental Results and Discussion}\label{sec5}
To assess the effectiveness of the proposed system, we designed an experimental setup grounded in real-world medical documentation. The evaluation was structured to emulate a realistic use case in which the system must interpret and respond to domain-specific queries related to vaccination guidelines.\\
Furthermore, to ensure the evaluation was contextually relevant, a custom question-answering (QA) benchmark was developed. This benchmark was tailored to the medical and procedural nuances of vaccination protocols, enabling a focused assessment of the system's retrieval and reasoning capabilities in a practical healthcare setting.

\subsection{Vaccination-Guide QA Benchmark}
To construct a domain-specific QA dataset, we employed an Automated Question Generation (AQG) approach. AQG is a Natural Language Processing (NLP) technique that generates questions from textual inputs by leveraging computational linguistics, cognitive theory, and machine learning. The question design is informed by the Question-Answer Relationship (QAR) framework, ensuring alignment with various cognitive levels of understanding.
Question difficulty is determined using a heuristic function grounded in Bloom’s Taxonomy:

\begin{itemize}
\item \textbf{Factual (Easy):} Focused on direct recall of information.
\item \textbf{Conceptual (Medium):} Requires synthesis and comprehension.
\item \textbf{Applied (Hard):} Involves reasoning and hypothetical scenarios.
\end{itemize}

To manage computational complexity and ensure full content coverage, the source document is segmented into smaller chunks. Each chunk is independently processed to generate three questions, resulting in a diverse and comprehensive QA dataset.

\begin{figure}[h!]
\centering
\includegraphics[width=0.95\textwidth]{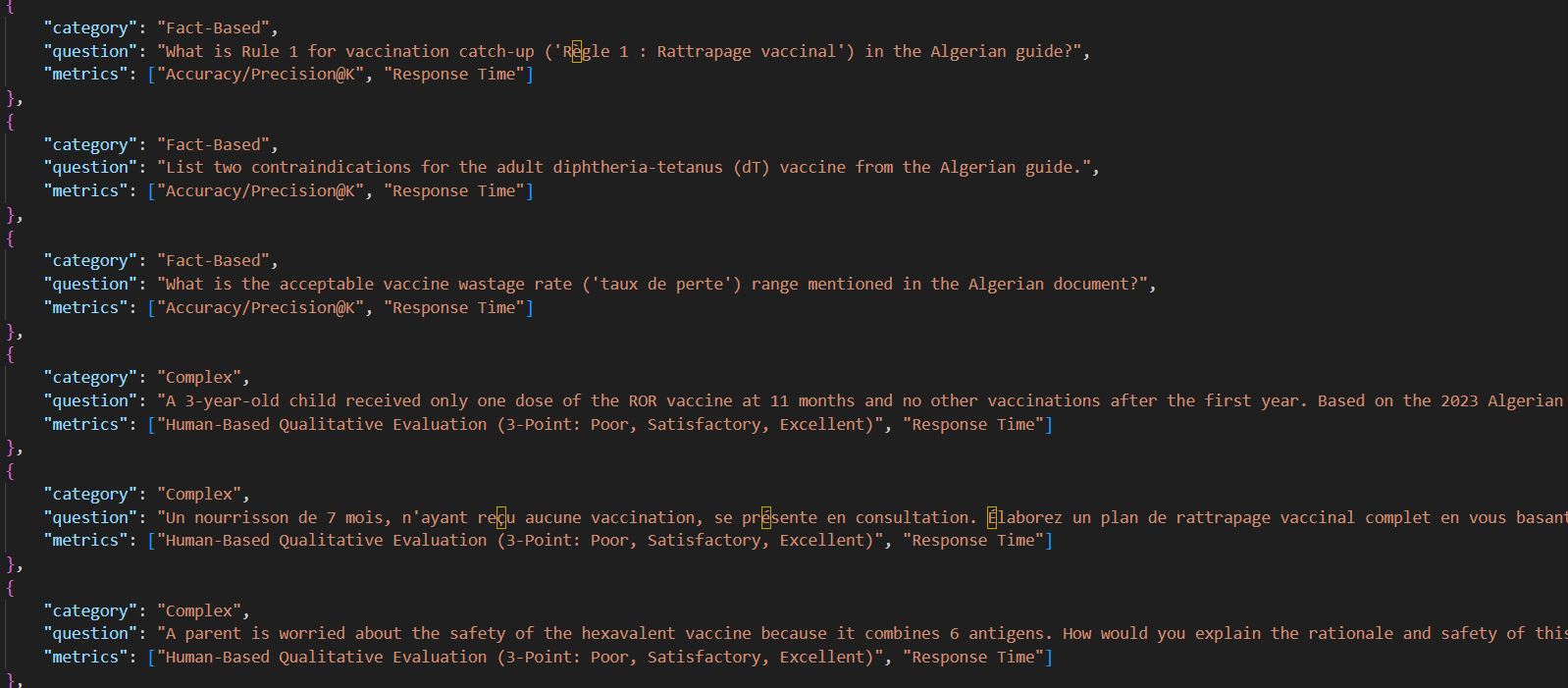}
\caption{Example entry from the vaccination QA benchmark dataset.}
\label{fig:Dataset}
\end{figure}

The AQG pipeline was designed with the following objectives:
\begin{enumerate}
\item Generate diverse questions from the French-language vaccination guide.
\item Categorize questions by type and difficulty level.
\item Output a structured JSON dataset suitable for LLM fine-tuning and evaluation.
\end{enumerate}
A sample from the generated QA dataset is shown in Figure~\ref{fig:Dataset}.

The implementation utilizes the \textbf{Google Gemini API} (gemini-2.0-flash) via the \texttt{langchain-google-genai} interface for prompt-based question generation. The system is served using the \textbf{FastAPI} framework to manage request handling, task monitoring, and dataset retrieval.

\subsection{Baseline Systems}
To assess the performance and relative effectiveness of different retrieval and generation strategies, we implemented four baseline systems, each representing a distinct approach within the RAG and LLM integration spectrum.
\begin{itemize}
\item \textbf{Simple RAG:}
A baseline Retrieval-Augmented Generation system that utilizes semantic vector embeddings for retrieving relevant document chunks. It serves as a foundational benchmark for evaluating semantic retrieval performance without additional enhancements.
\item \textbf{Enhanced RAG:}  
An improved RAG pipeline incorporating advanced preprocessing techniques, including structural parsing, noise filtering, and dynamic context windowing. These refinements aim to improve the relevance and granularity of retrieved information.
\item \textbf{Fine-Tuned Mistral-7B:}  
A large multilingual language model (Mistral-7B) fine-tuned on a curated dataset of vaccination-related texts. This model is trained to understand domain-specific terminology better and respond with higher accuracy in French and Arabic.

\item \textbf{Agent-Based System:}  
A more sophisticated architecture integrating an agentic reasoning layer. This system employs autonomous agents capable of decomposing user queries, selecting relevant tools (e.g., document retrievers or reasoning modules), and orchestrating multi-step responses tailored to complex medical inquiries.
\end{itemize}
These baseline systems enable a comparative analysis of trade-offs between simplicity, domain adaptation, and reasoning capabilities in multilingual medical QA settings.

\subsection{Implementation Details}
The development of the baseline systems relied on state-of-the-art tools for document parsing, semantic retrieval, and natural language understanding, with particular emphasis on French-language processing.

\textbf{Simple RAG:}
The initial RAG-based system was implemented using standard document parsing tools alongside natural language processing techniques tailored for French-language content. Semantic retrieval was enabled through a French embedding model obtained from HuggingFace, with vector representations stored in FAISS (Facebook AI Similarity Search). The Mistral-7B language model was used for generation, accessed via the HuggingFace Transformers library. To accommodate computational limitations in environments such as Google Colab, quantization techniques were applied. LangChain served as the orchestration framework for managing retrieval and generation tasks. The system was designed to process structured vaccination documents and respond to user queries by distinguishing between mandatory and recommended vaccines, while also highlighting any missing contextual information.

\textbf{Enhanced RAG:}
The second system focused on improving retrieval accuracy and semantic coverage. Instead of using naive chunking, advanced document parsing was performed using the Unstructured library. A multilingual embedding model from HuggingFace was utilized to represent domain-specific terminology more effectively. The resulting embeddings were stored using FAISS to enable efficient similarity search. Retrieved content was then passed to the Gemini model for answer generation. This system, also deployed in Google Colab, demonstrated improved performance in terms of accuracy and response quality over the simple RAG baseline, without increasing resource demands.

\textbf{Agent-Based RAG:}
To handle more complex and multi-step user queries, an agent-based system was developed using the \textit{LlamaIndex} framework. This system utilized the \textit{ReActAgent}, which enables structured reasoning by decomposing questions into sequential tasks. A suite of retrieval tools was defined, each corresponding to specific sources, including the Algerian national vaccination guide and WHO guidelines. Based on the user query, the agent selected the most appropriate tools and navigated to their contents accordingly.
Semantic representations were generated using the \textit{HuggingFaceEmbedding} module, with document vectors stored in \textit{ChromaDB}. To support hybrid retrieval, BM25 keyword-based ranking was integrated via the \textit{rank\_bm25} library. Generation and reasoning were handled using Google's Gemini model, accessed through the \textit{llama-index-llms-google-genai} connector. This configuration enabled coherent, context-aware responses, especially for fact-intensive or multi-source questions.

\textbf{Fine-Tuned Mistral-7B:}
The final approach involved domain adaptation and task-specific fine-tuning of the \textit{mistralai/Mistral-7B-v0.1} model. This pipeline utilized Quantized Low-Rank Adaptation (QLoRA) to enhance training efficiency and minimize memory consumption. Implementation was carried out using the Hugging Face ecosystem, incorporating \textit{transformers}, \textit{datasets}, \textit{peft}, \textit{bitsandbytes}, and \textit{accelerate}, with \textit{PyTorch} as the foundational framework.

The fine-tuning process was conducted in two stages: (1) continued pre-training on a French-language vaccination guide to adapt the model to domain-specific context, and (2) supervised fine-tuning using a question–answer dataset to enhance QA capabilities. All training and inference tasks were performed in a GPU-enabled Google Colab environment using 4-bit quantization and parameter-efficient fine-tuning (PEFT). The result was a lightweight LoRA adapter, which, when combined with the base Mistral model, enabled the effective and domain-specific generation of responses for vaccination-related queries.

\subsection{Performance Metrics and Evaluation Criteria}
To rigorously assess the effectiveness and reliability of the proposed systems, we adopted a multifaceted evaluation strategy that combined expert validation, human qualitative assessment, and system performance timing metrics.

\textbf{Accuracy:}
The primary metric for system performance was factual accuracy, determined by domain experts. Each answer was manually evaluated against the official Algerian vaccination guidelines. This metric extended beyond textual consistency with source material, focusing on the system's ability to interpret and apply medical guidelines correctly. This ensured that outputs were not only accurate but also clinically meaningful and trustworthy.
\[
\text{Accuracy} = \frac{\text{Number of Correct Answers}}{\text{Total Number of Questions}}
\]

\textbf{Human-Based Qualitative Evaluation:}
To complement factual accuracy, a qualitative assessment was conducted using a three-point Likert scale, where responses were rated as follows:
\begin{itemize}
\item \textbf{Excellent (2)} – Fully correct, comprehensive, and clearly articulated responses.
\item \textbf{Satisfactory (1)} – Generally accurate with minor omissions or slight ambiguity.
\item \textbf{Poor (0)} – Inaccurate, misleading, or insufficient for clinical use.
\end{itemize}

Each model response was scored accordingly, and an aggregate human evaluation score was computed:
\[
\text{Mean Human Score} = \frac{1}{N} \sum_{i=1}^{N} s_i \quad \text{where } s_i \in \{0, 1, 2\}
\]

\textbf{Response Time:}
To assess the practical viability of each system, response latency was measured. This metric captured the total time taken from query submission to response generation:
\[
\text{Average Response Time} = \frac{1}{N} \sum_{i=1}^{N} t_i \quad \text{where } t_i \text{ is time in seconds for question } i
\]
This helped identify trade-offs between reasoning complexity and computational efficiency, particularly relevant in time-sensitive clinical environments.

\textbf{Qualitative Breakdown for Complex Queries:}
To evaluate system robustness on high-difficulty queries, human-rated responses were further analyzed for distribution across quality categories. The percentage breakdown for complex queries is defined as:

Let \( N_{\text{excellent}} \), \( N_{\text{satisfactory}} \), and \( N_{\text{poor}} \) be the number of responses classified into each category. Let \( N_{\text{complex}} \) be the total number of complex questions.

\[
\text{Excellent (\%)} = \left( \frac{N_{\text{excellent}}}{N_{\text{complex}}} \right) \times 100
\]

\[
\text{Satisfactory (\%)} = \left( \frac{N_{\text{satisfactory}}}{N_{\text{complex}}} \right) \times 100
\]

\[
\text{Poor (\%)} = \left( \frac{N_{\text{poor}}}{N_{\text{complex}}} \right) \times 100
\]

This analysis enabled a detailed understanding of how each system handled reasoning-intensive questions, which are typically more representative of real-world medical use cases.

\subsection{Comparative Analysis}
To assess the performance of the proposed systems, namely, the fine-tuned language model, the agent-based retrieval system, and the enhanced RAG pipeline, we conducted evaluations using three categories of questions, each designed to represent different cognitive complexities and real-world use cases.

Table~\ref{tab:overall_metrics} summarizes the overall performance metrics averaged across all question types. Among the systems evaluated, AgenticRAG demonstrated the highest average accuracy, though this was accompanied by increased response latency. Both AgenticRAG and EnhancedRAG achieved a perfect citation rate, ensuring the traceability and verifiability of generated responses. In contrast, SimpleRAG, serving as the baseline system, yielded considerably lower accuracy despite similar or higher response times. The fine-tuned model exhibited no response capability in this evaluation setup, likely due to limitations in generalization or incomplete domain adaptation.

\begin{table}[ht]
\caption{Summary of System Performance Across All Question Types}
\label{tab:overall_metrics}
\centering
\begin{tabular}{lccc}
\toprule
\textbf{System} & \textbf{Average Score} & \textbf{Avg. Response Time (s)} & \textbf{Citation Rate (\%)} \\
\midrule
AgenticRAG        & 0.73 & 16.72 & 100 \\
EnhancedRAG       & 0.43 & 5.10  & 100 \\
SimpleRAG         & 0.03 & 20.48 & 100 \\
Fine-Tuned Model  & 0.00 & 0.00  & 0   \\
\bottomrule
\end{tabular}
\end{table}

\subsubsection{Detailed Performance by Question Category}
To gain deeper insights into system capabilities, we analyzed performance across three distinct question categories: fact-based, complex reasoning, and cross-document queries. The detailed results are presented in Table~\ref{tab:category_metrics}.

AgenticRAG achieved the highest accuracy in both complex and cross-document question types, underscoring its strength in handling multi-step reasoning and content aggregation from disparate sources. EnhancedRAG demonstrated relatively better performance on fact-based questions and maintained consistently low response times across all categories, indicating its efficiency and suitability for real-time applications. Conversely, SimpleRAG exhibited poor accuracy and variable latency, further validating its role as a minimal baseline. The fine-tuned model failed to produce valid outputs for any category under the evaluation setup, likely due to insufficient domain alignment or generalization.

\begin{table}[ht]
\caption{System Performance by Question Category}
\label{tab:category_metrics}
\centering
\begin{tabular}{lccc}
\toprule
\textbf{Metric} & \textbf{Fact-Based} & \textbf{Complex} & \textbf{Cross-Document} \\
\midrule
Accuracy (AgenticRAG)   & 0.50 & 1.00 & 0.70 \\
Accuracy (EnhancedRAG)  & 0.70 & 0.60 & 0.00 \\
Accuracy (SimpleRAG)    & 0.10 & 0.00 & 0.00 \\
Accuracy (Fine-Tuned)   & 0.00 & 0.00 & 0.00 \\
Response Time (s) AgenticRAG   & 22.20 & 12.09 & 15.87 \\
Response Time (s) EnhancedRAG  & 4.60  & 6.27  & 4.41 \\
Response Time (s) SimpleRAG    & 13.48 & 32.57 & 15.40 \\
\bottomrule
\end{tabular}
\end{table}

\subsubsection{Qualitative Evaluation of Complex Scenarios}
For complex scenario questions, a qualitative evaluation was conducted by categorizing system responses into three levels: \textit{Excellent} (fully correct and contextually appropriate), \textit{Satisfactory} (partially correct or moderately helpful), and \textit{Poor} (irrelevant, incorrect, or misleading). The distribution of these assessments is summarized in Table~\ref{tab:qualitative_complex}.

AgenticRAG achieved the most balanced performance, with 40\% of its responses rated as Excellent and an additional 20\% deemed Satisfactory. This suggests a high degree of contextual reasoning and synthesis, enabled by its agentic design. In contrast, EnhancedRAG generated a significant number of Poor responses (70\%), with only 30\% marked Excellent and none rated Satisfactory, indicating inconsistency in handling complex questions.

The SimpleRAG system failed to produce any acceptable responses during this evaluation, with all outputs classified as 'Poor'. Similarly, the Fine-Tuned Model yielded no usable responses, underscoring the challenges of domain-specific fine-tuning without extensive and high-quality task supervision.

\begin{table}[ht]
\centering
\caption{Qualitative Breakdown for Complex Questions}
\label{tab:qualitative_complex}
\begin{tabular}{lccc}
\toprule
\textbf{System} & \textbf{Excellent (\%)} & \textbf{Satisfactory (\%)} & \textbf{Poor (\%)} \\
\midrule
AgenticRAG        & 40 & 20 & 40 \\
EnhancedRAG       & 30 & 0  & 70 \\
SimpleRAG         & 0  & 0  & 100 \\
Fine-Tuned Model  & 0  & 0  & 0 \\
\bottomrule
\end{tabular}
\end{table}

This analysis offers several key insights. AgenticRAG outperforms other systems in complex, context-rich scenarios, indicating its agent-based reasoning framework effectively enhances interpretability and coherence. However, this comes with increased computational overhead, as reflected in longer average response times.

EnhancedRAG, while less effective on nuanced queries, offers faster performance and greater reliability in straightforward tasks. This makes it more suitable for latency-sensitive or resource-constrained deployments.

SimpleRAG consistently underperforms in both accuracy and qualitative usefulness, reaffirming its role as a fundamental baseline. The Fine-Tuned Model's inability to generate coherent outputs further highlights the limitations of current low-resource fine-tuning strategies in specialized domains.

Notably, all systems that returned outputs maintained consistent source citation behavior—a critical feature for transparency and user trust in high-stakes domains such as healthcare.

\begin{figure}[ht]
    \centering
    \includegraphics[width=0.95\columnwidth]{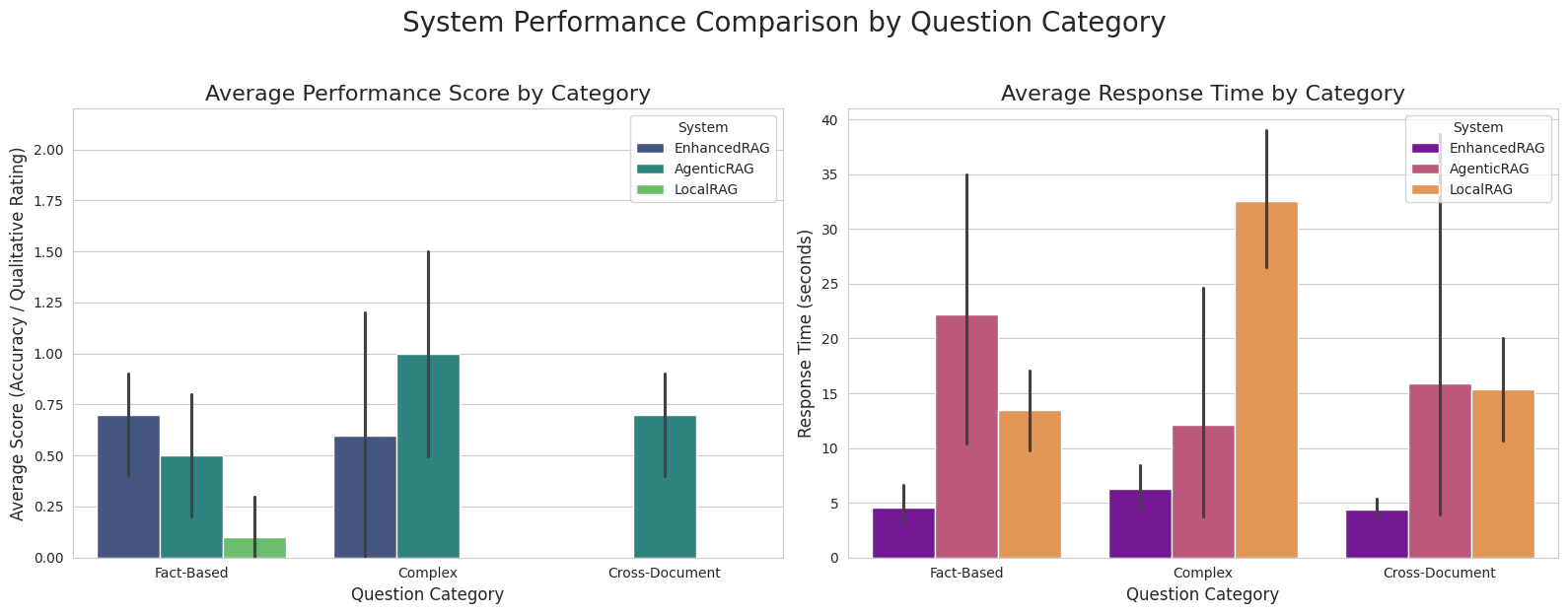}
    \caption{System Performance Comparison by Question Category.}
    \label{fig:performance_by_category}
\end{figure}

\begin{figure}[ht]
    \centering
    \includegraphics[width=0.95\columnwidth]{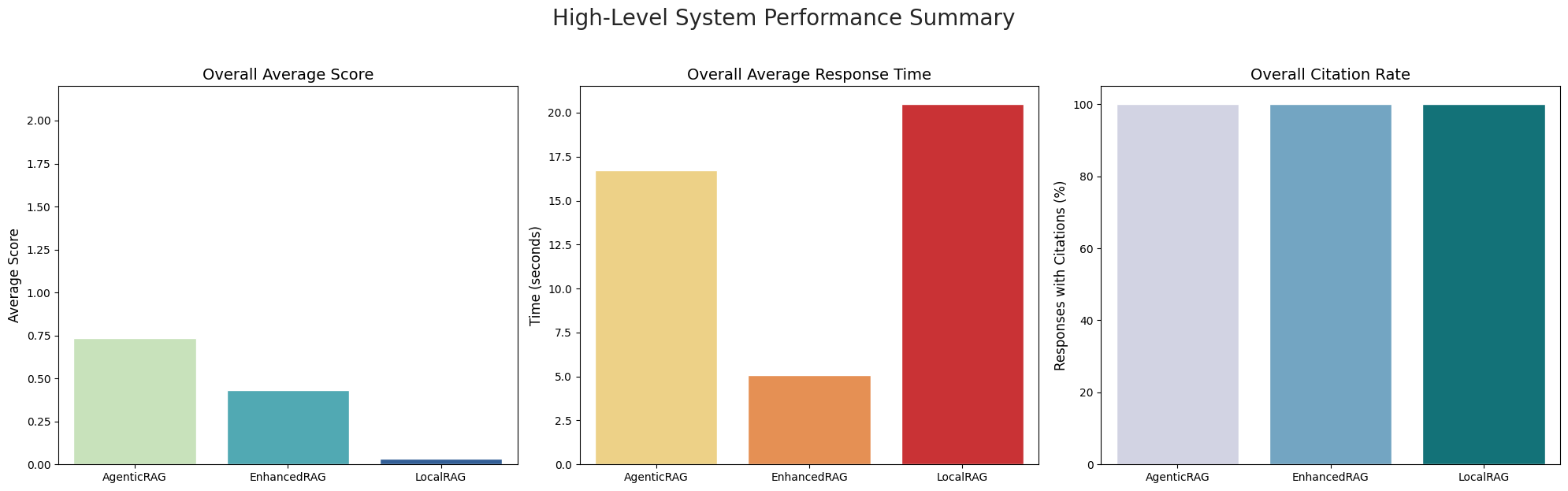}
    \caption{High-Level System Performance Summary.}
    \label{fig:overall_performance_summary}
\end{figure}

Figures~\ref{fig:performance_by_category} and \ref{fig:overall_performance_summary} provide visual summaries of performance across question categories and overall metrics.

\subsection{Discussion}
This project explored several approaches to building a robust question-answering (QA) system tailored to the Algerian vaccination program and associated WHO guidelines. Each method—ranging from fine-tuning to progressively more sophisticated RAG systems—highlighted distinct trade-offs in performance, scalability, and clinical reliability.

The initial strategy involved fine-tuning a language model on a small, domain-specific dataset. While this approach aimed to internalize relevant medical knowledge, it proved largely ineffective. The fine-tuned model frequently generated vague, inaccurate, or hallucinated responses, failing to deliver useful answers even on relatively simple queries. Its inability to handle contextual reasoning or multi-step synthesis tasks revealed the limitations of using fine-tuning with limited, narrowly scoped data. These results confirm that fine-tuning, while powerful in high-resource settings, is impractical and unreliable for low-resource clinical domains without access to large-scale annotated corpora.

In response to these challenges, the project pivoted to RAG, which grounds answers in dynamically retrieved source content. The first implementation, SimpleRAG, employed basic semantic search and direct context injection. While conceptually straightforward, this method struggled with depth, often returning incomplete or imprecise answers. Nevertheless, it served as a useful baseline for evaluating subsequent improvements.

EnhancedRAG was built on this foundation by improving the retrieval pipeline and refining the context structure before generation. This system delivered more accurate and faster responses, particularly for fact-based queries. Its efficiency and high citation consistency made it a strong candidate for time-sensitive applications. However, EnhancedRAG’s single-pass generation framework limited its ability to handle more complex or cross-referenced questions.

To address these limitations, the final system—AgenticRAG—introduced a modular, agent-based reasoning structure. By dividing the QA task into sequential, specialized steps (query reformulation, iterative retrieval, and answer synthesis), AgenticRAG was able to reason more effectively across multiple document sections. This design significantly improved performance on complex and cross-document questions. Although slower than EnhancedRAG, its answers were more accurate, complete, and clinically reliable.

Overall, the iterative development process revealed a clear trajectory of increasing performance: the fine-tuned model was quickly deemed ineffective, SimpleRAG served as a minimal baseline, EnhancedRAG struck a balance between speed and factual precision, and AgenticRAG ultimately provided the most reliable and robust outputs, particularly for challenging queries. These findings suggest that, in constrained domains with limited training data, agent-based RAG systems offer the most promising path forward for trustworthy and high-utility clinical QA.

\section{Proposed Mobile Application for Vaccination Chatbot Interaction}
This section introduces the proposed mobile application designed to facilitate efficient, multilingual, and evidence-based interaction between doctors and a vaccination-focused AI chatbot. The application serves as a user-friendly platform enabling Algerian healthcare professionals to access verified vaccination information directly from trusted national and international sources.

Upon secure login, doctors can initiate or revisit chat sessions with the AI assistant. They may pose questions in Arabic, French, or English, and the system responds with answers grounded in Algeria’s national immunization guide, World Health Organization (WHO) guidelines, and other approved documents. Each response includes a direct citation, pointing to the exact page or paragraph from which the information was extracted, enabling on-the-spot verification and fostering transparency.
\begin{figure}[h!]
	\centering
	\begin{minipage}[b]{0.21\textwidth}
		\centering
		\includegraphics[width=\textwidth]{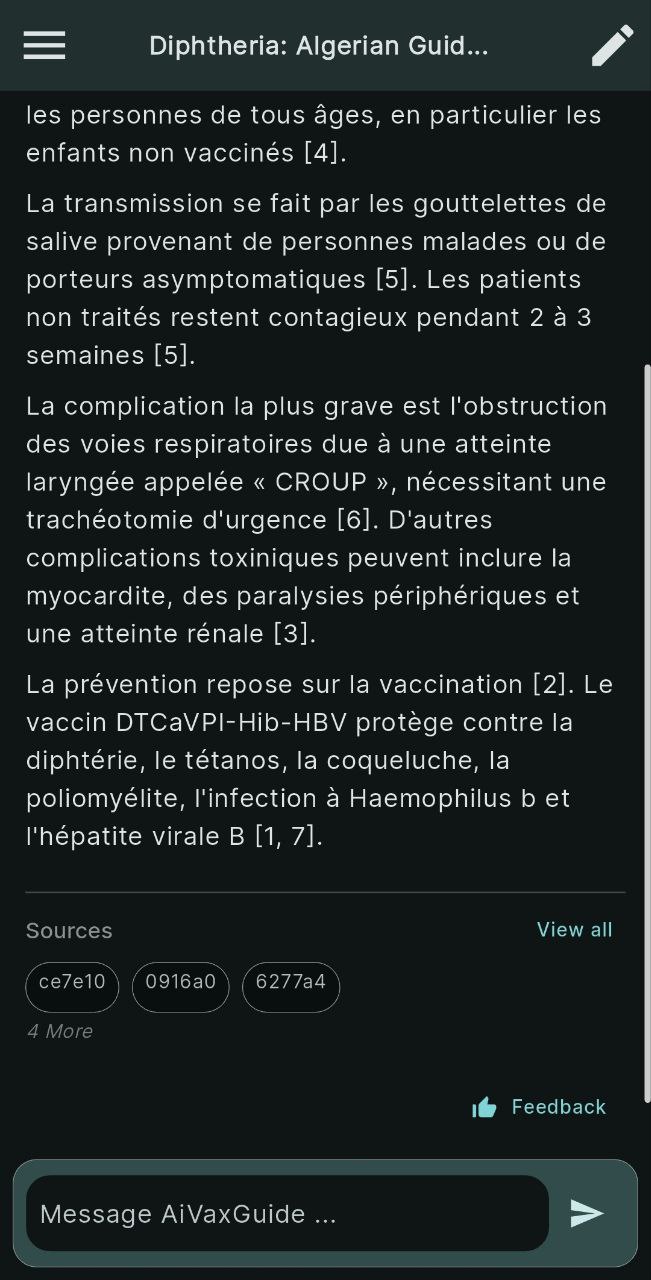}
		\label{fig:sendmsg}
	\end{minipage}
	\hspace{0.006\textwidth} 
	\begin{minipage}[b]{0.21\textwidth}
		\centering
		\includegraphics[width=\textwidth]{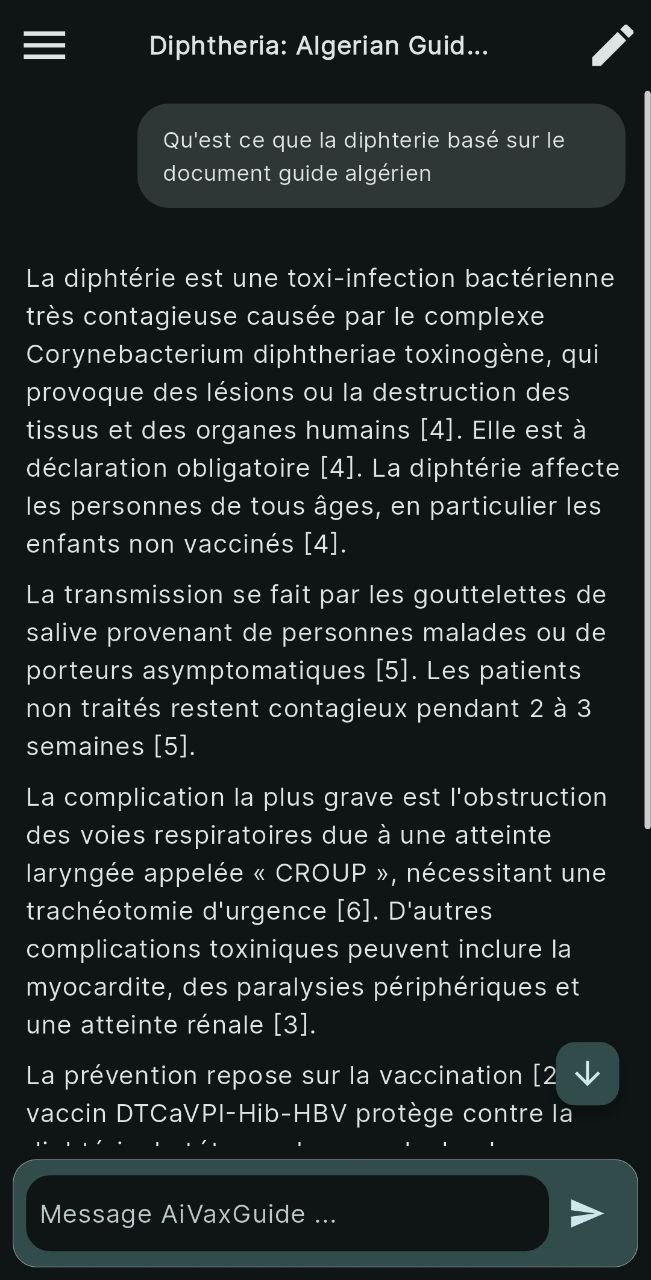}
		\label{fig:sendmsg2}
	\end{minipage}
	\caption{Chatbot interaction showing (left) a continuation of the doctor's message and (right) the initial message being sent.}
	\label{fig:chat_messages}
\end{figure}

The chat interface is designed to resemble familiar messaging platforms, as illustrated in Figure~\ref{fig:chat_messages}, providing a smooth and intuitive user experience. Doctors can maintain multiple conversations simultaneously, allowing for case-specific or topic-specific organization. The chatbot supports two distinct operational modes—agentic and enhanced—providing flexibility according to the user's needs. Conversations are automatically saved, allowing doctors to revisit prior exchanges and continue discussions seamlessly.
A key functionality of the system is intelligent question answering, where doctors can inquire about vaccine schedules, contraindications, side effects, or any immunization-related concern. As shown in Figure~\ref{fig:chat_messages}, the chatbot delivers concise responses, backed by citations from the original source material.

\begin{figure}[t!]
	\centering
	\begin{minipage}[b]{0.21\textwidth}
		\centering
		\includegraphics[width=\textwidth]{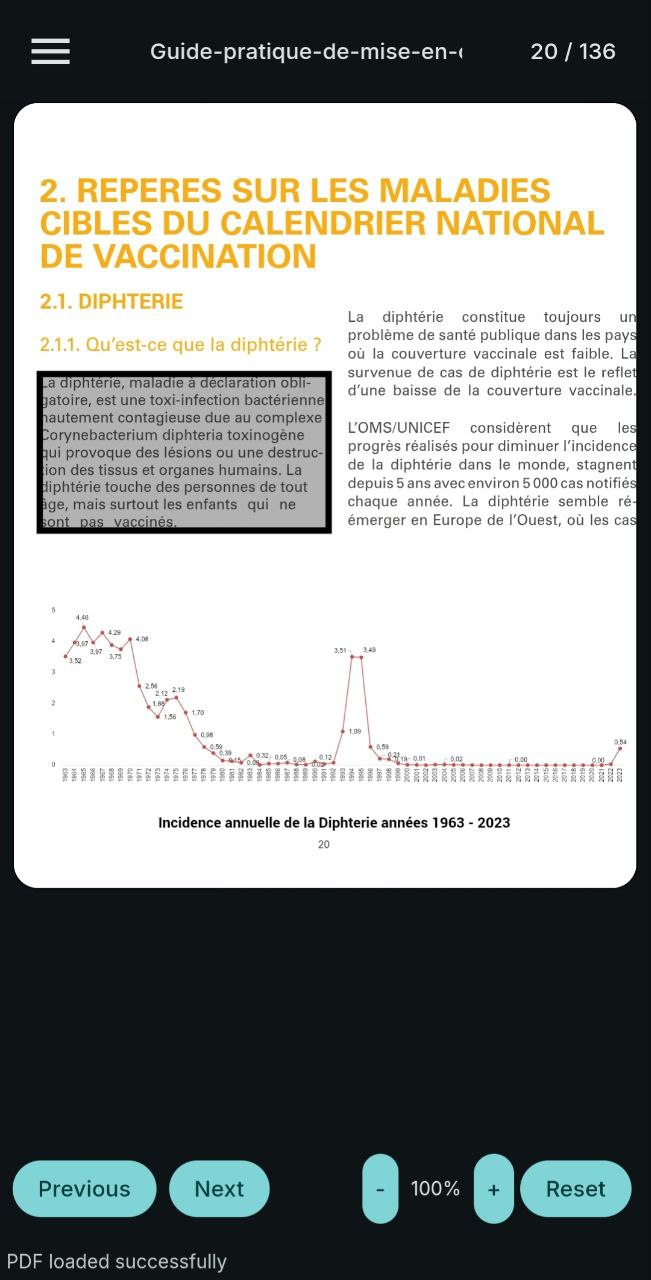}
		\label{fig:source}
	\end{minipage}
	\hspace{0.006\textwidth} 
	\begin{minipage}[b]{0.21\textwidth}
		\centering
		\includegraphics[width=\textwidth]{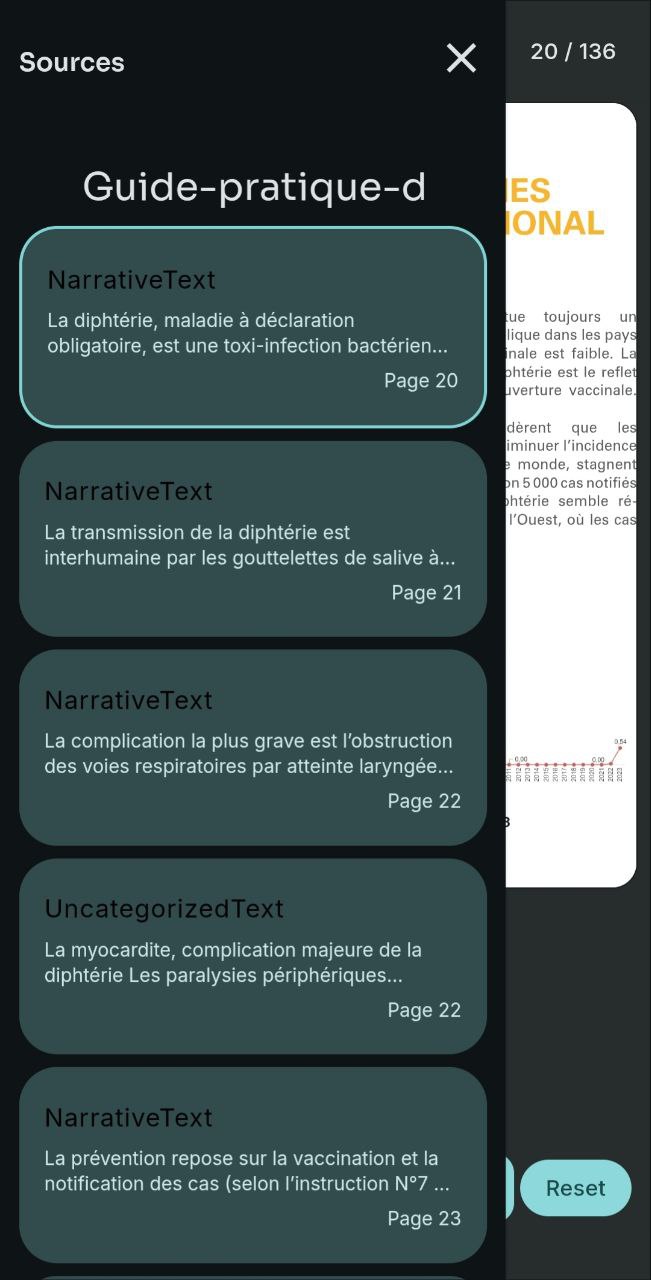}
		\label{fig:relevant}
	\end{minipage}
	\caption{Visual explanation of the source citation mechanism: (left) the document from which the answer is retrieved and (right) the highlighted relevant content supporting the chatbot's response.}
	\label{fig:source_highlight}
\end{figure}

To enhance information reliability, the system highlights the specific segments from which answers are drawn. This feature allows users to view the relevant excerpts directly within the original documents, as demonstrated in Figure~\ref{fig:source_highlight}. This approach ensures traceability and builds confidence in the accuracy of the chatbot’s responses.

\begin{figure}[h!]
	\centering
	\includegraphics[width=0.21\textwidth]{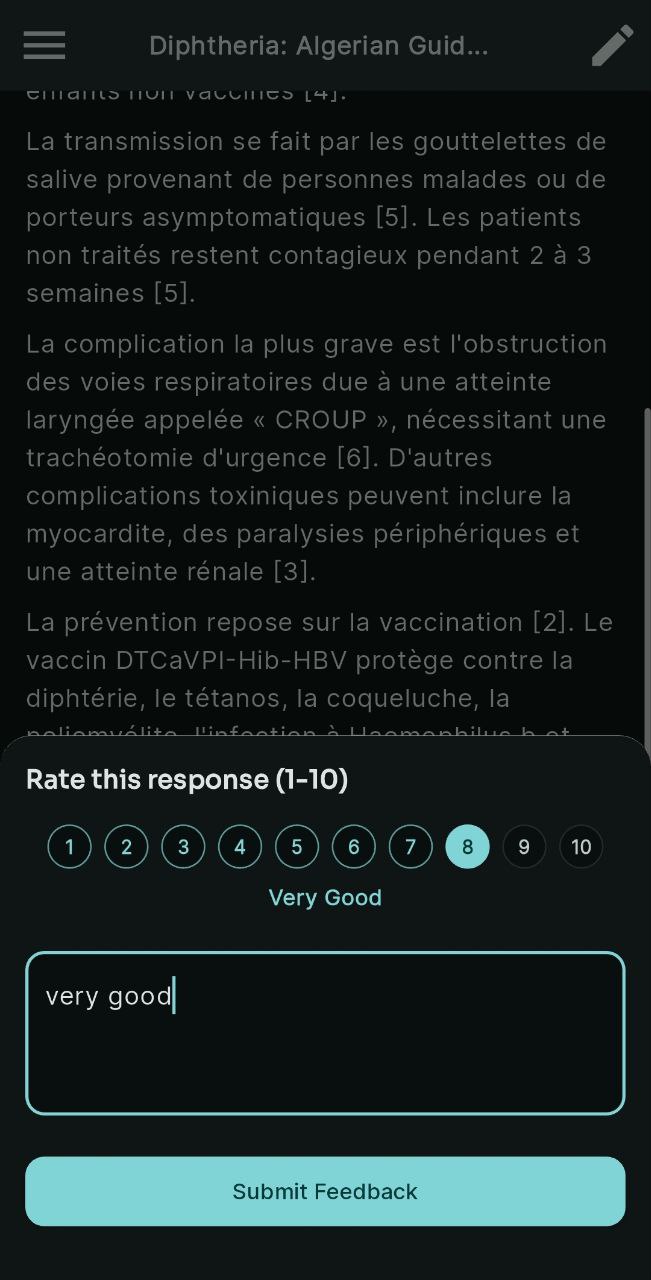}
	\caption{Rating feature.}
	\label{fig:rating}
\end{figure}

In support of continuous improvement, the application incorporates a feedback and rating mechanism. After each interaction, doctors are invited to evaluate the chatbot’s response on a scale from 0 to 10 and optionally provide comments. This feedback loop is vital for identifying issues, refining performance, and aligning the chatbot’s responses more closely with users’ expectations. An example of this functionality is depicted in Figure~\ref{fig:rating}.

\section{Conclusion and Future Work}\label{sec6}
This work presents a structured approach to improving access to immunization guidelines through the application of LLMs, supporting clinical decision-making in resource-constrained settings. The study began with a comparative analysis of national (Algerian) and international (WHO) vaccination protocols, utilizing statistical and Natural Language Processing techniques to establish a foundational understanding of content variability and complexity across sources.

Building upon this foundation, we designed and iteratively improved a question-answering (QA) system that progressed from a baseline RAG model to a more capable agent-based reasoning system, referred to as AgenticRAG. This final system integrates query reformulation, iterative retrieval, and structured response generation, resulting in significantly enhanced performance on complex and ambiguous clinical queries. Deployed via a user-friendly mobile application, the system provides clinicians with reliable and context-aware responses, grounded in authoritative medical documents.

Quantitative and qualitative evaluations confirmed that AgenticRAG outperformed simpler RAG configurations and fine-tuned LLMs across multiple performance dimensions, including accuracy, contextual relevance, and handling of cross-document reasoning tasks. Notably, its ability to synthesize multi-step answers demonstrated its potential as a practical decision-support tool in clinical workflows.

A key challenge encountered during development was the handling of sensitive and structurally complex medical documents. These often required manual curation to ensure reliable preprocessing and avoid misinterpretation by automated tools. Additionally, current LLMs, even when enhanced with retrieval or agentic mechanisms, still exhibit limitations in reasoning through highly specialized or nuanced clinical logic, emphasizing the need for domain-specific adaptation.

Future work will focus on three primary directions. First, incorporating larger and more diverse datasets, including dynamically updated guidelines from health authorities, will improve the breadth and currency of the system’s knowledge base. Second, enabling multilingual support will expand accessibility for healthcare providers operating in linguistically diverse regions. Ultimately, refining the agent-based architecture to incorporate more advanced planning and decision-making strategies may narrow the gap between automated and expert-level reasoning, thereby further enhancing the utility of LLMs in complex medical domains.

\section*{Declaration on Generative AI}
During the preparation of this work, the authors used ChatGPT and Grammarly to rephrase and perform Grammar and spelling checks. After using these tools, the authors reviewed and edited the content as needed. The authors take full responsibility for the publication’s content.

\bibliographystyle{unsrt}  
\bibliography{AIVAXGUIDE}

\end{document}